\pdfoutput=1

\documentclass[runningheads]{llncs}

\usepackage{accv}
\usepackage{accvabbrv}                 %

\usepackage{lmodern}
\usepackage{graphicx}
\usepackage{booktabs}
\usepackage{multirow}
\usepackage{makecell}
\usepackage{microtype}
\usepackage{placeins}
\usepackage{float}
\usepackage{afterpage}
\usepackage{marvosym}  %
\usepackage{orcidlink}
\NewCommandCopy{\origorcidlink}{\orcidlink}\RenewDocumentCommand{\orcidlink}{m}{{\hypersetup{pdfborder={0 0 0}}\origorcidlink{#1}}}
\makeatletter
\renewcommand\subsection{\@startsection{subsection}{2}{\z@}%
                       {-10\p@ \@plus -3\p@ \@minus -3\p@}%
                       {4\p@ \@plus 2\p@ \@minus 2\p@}%
                       {\normalfont\normalsize\bfseries\boldmath
                        \rightskip=\z@ \@plus 8em\pretolerance=10000 }}
\makeatother
\usepackage{tikz}
\usetikzlibrary{arrows.meta,positioning,calc,backgrounds,fit,shapes.geometric}

\usepackage{hyperref}

\newif\ifcameraready %

\newcommand{\cmark}{\checkmark}

\newcommand{\pix}{\mathbf{u}}                    %
\newcommand{\ptmap}{\mathbf{X}}                  %
\newcommand{\tu}{\mathbf{t}_{x}}                 %
\newcommand{\tv}{\mathbf{t}_{y}}                 %
\newcommand{\nhat}{\hat{\mathbf{n}}}             %
\newcommand{\eone}{\hat{\mathbf{e}}_{1}}         %
\newcommand{\etwo}{\hat{\mathbf{e}}_{2}}         %
\newcommand{\rdir}{\mathbf{d}}                   %
\newcommand{\cam}{\mathbf{R}_{\mathrm{c2w}}}     %

\newcommand{\Rana}{\mathbf{R}_{\mathrm{a}}}      %
\newcommand{\Rres}{\mathbf{R}_{\Delta}}          %
\newcommand{\Rfin}{\mathbf{R}}                   %
\newcommand{\qana}{q_{\mathrm{a}}}               %
\newcommand{\dq}{\Delta q}                       %
\newcommand{\qfin}{q}                            %

\newcommand{\sbarone}{\bar{s}_{1}}               %
\newcommand{\sbartwo}{\bar{s}_{2}}               %
\newcommand{\Scov}{\boldsymbol{\Sigma}_{\mathrm{s}}}   %
\newcommand{\Sfin}{\boldsymbol{\Sigma}}                %

\newcommand{\nScenes}{6{,}474}

\newcommand{\runtimeOverheadPct}{2\%}

\newcommand{\valBaseLargePSNR}{27.519}
\newcommand{\valBaseLargeSSIM}{0.892}
\newcommand{\valBaseLargeLPIPS}{0.112}

\newcommand{\valCoreLargePSNR}{27.686}
\newcommand{\valCoreLargeSSIM}{0.893}
\newcommand{\valCoreLargeLPIPS}{0.109}

\newcommand{\valBaseLargeHRRCfourPSNR}{24.863}
\newcommand{\valBaseLargeHRRCfourSSIM}{0.842}
\newcommand{\valBaseLargeHRRCfourLPIPS}{0.316}
\newcommand{\valBaseLargeHRRCdropPSNR}{2.656}

\newcommand{\valBaseLargeHRRCtwoPSNR}{26.366}
\newcommand{\valBaseLargeHRRCtwoSSIM}{0.868}
\newcommand{\valBaseLargeHRRCtwoLPIPS}{0.205}

\newcommand{\valCoreLargeHRRCfourPSNR}{25.182}
\newcommand{\valCoreLargeHRRCfourSSIM}{0.848}
\newcommand{\valCoreLargeHRRCfourLPIPS}{0.296}
\newcommand{\valCoreLargeHRRCdropPSNR}{2.504}

\newcommand{\valCoreLargeHRRCtwoPSNR}{26.493}
\newcommand{\valCoreLargeHRRCtwoSSIM}{0.869}
\newcommand{\valCoreLargeHRRCtwoLPIPS}{0.198}

\newcommand{\dqAngleLargeLate}{0.160}
\newcommand{\dqAngleLargeLateDeg}{9.1}
\newcommand{\dqPninetynineLargeLate}{1.073}
\newcommand{\dqPninetynineLargeLateDeg}{61.5}
\newcommand{\frameOrthErrLargeLate}{1.2\times10^{-4}}
\newcommand{\frameOrthErrBaseLargeLate}{5.4\times10^{-3}}
\newcommand{\frameDetMinLargeLate}{1.000}
\newcommand{\frameDetMinBaseLargeLate}{0.951}
\newcommand{\anisoCoreLargeLate}{0.585}
\newcommand{\anisoBaseLargeLate}{0.667}
\newcommand{\isoFracCoreLargeLate}{0.134}
\newcommand{\isoFracBaseLargeLate}{0.209}
\newcommand{\scalesMulMinFracCoreLargeLate}{0.364}
\newcommand{\scalesMulMinFracBaseLargeLate}{0.605}

\newcommand{\psWinPSNRLarge}{70\%}

\newcommand{\psWinSSIMLarge}{70\%}

\newcommand{\psWinLPIPSLarge}{81\%}

\newcommand{\psWinPSNRLargeFour}{62\%}

\newcommand{\psWinSSIMLargeFour}{57\%}

\newcommand{\psWinLPIPSLargeFour}{66\%}

\newcommand{\psWinPSNRLargeEight}{67\%}

\newcommand{\psWinSSIMLargeEight}{62\%}

\newcommand{\psWinLPIPSLargeEight}{66\%}

\newcommand{\valMvPSNR}{26.387}
\newcommand{\valMvSSIM}{0.869}
\newcommand{\valMvLPIPS}{0.128}
\newcommand{\valMvTwoPSNR}{20.294}
\newcommand{\valMvTwoSSIM}{0.808}
\newcommand{\valMvTwoLPIPS}{0.279}
\newcommand{\valMvFourPSNR}{17.882}
\newcommand{\valMvFourSSIM}{0.754}
\newcommand{\valMvFourLPIPS}{0.421}
\newcommand{\valMvEightPSNR}{17.195}
\newcommand{\valMvEightSSIM}{0.770}
\newcommand{\valMvEightLPIPS}{0.438}
\newcommand{\valTransPSNR}{26.688}
\newcommand{\valTransSSIM}{0.875}
\newcommand{\valTransLPIPS}{0.125}
\newcommand{\valTransTwoPSNR}{20.499}
\newcommand{\valTransTwoSSIM}{0.814}
\newcommand{\valTransTwoLPIPS}{0.275}
\newcommand{\valTransFourPSNR}{18.071}
\newcommand{\valTransFourSSIM}{0.760}
\newcommand{\valTransFourLPIPS}{0.423}
\newcommand{\valTransEightPSNR}{17.383}
\newcommand{\valTransEightSSIM}{0.776}
\newcommand{\valTransEightLPIPS}{0.438}
\newcommand{\valDepthPSNR}{27.506}
\newcommand{\valDepthSSIM}{0.890}
\newcommand{\valDepthLPIPS}{0.112}
\newcommand{\valDepthTwoPSNR}{19.914}
\newcommand{\valDepthTwoSSIM}{0.772}
\newcommand{\valDepthTwoLPIPS}{0.332}
\newcommand{\valDepthFourPSNR}{16.299}
\newcommand{\valDepthFourSSIM}{0.631}
\newcommand{\valDepthFourLPIPS}{0.491}
\newcommand{\valDepthEightPSNR}{15.079}
\newcommand{\valDepthEightSSIM}{0.619}
\newcommand{\valDepthEightLPIPS}{0.508}
\newcommand{\valHiPSNR}{27.194}
\newcommand{\valHiSSIM}{0.882}
\newcommand{\valHiLPIPS}{0.117}
\newcommand{\valHiTwoPSNR}{25.097}
\newcommand{\valHiTwoSSIM}{0.867}
\newcommand{\valHiTwoLPIPS}{0.191}
\newcommand{\valHiFourPSNR}{24.097}
\newcommand{\valHiFourSSIM}{0.876}
\newcommand{\valHiFourLPIPS}{0.244}
\newcommand{\valHiEightPSNR}{23.816}
\newcommand{\valHiEightSSIM}{0.910}
\newcommand{\valHiEightLPIPS}{0.248}
\newcommand{\valPixPSNR}{26.073}
\newcommand{\valPixSSIM}{0.864}
\newcommand{\valPixLPIPS}{0.136}
\newcommand{\valPixTwoPSNR}{25.262}
\newcommand{\valPixTwoSSIM}{0.855}
\newcommand{\valPixTwoLPIPS}{0.208}
\newcommand{\valPixFourPSNR}{24.378}
\newcommand{\valPixFourSSIM}{0.864}
\newcommand{\valPixFourLPIPS}{0.275}
\newcommand{\valPixEightPSNR}{24.003}
\newcommand{\valPixEightSSIM}{0.898}
\newcommand{\valPixEightLPIPS}{0.280}
\newcommand{\valBaseLargeHRRCeightPSNR}{23.342}
\newcommand{\valBaseLargeHRRCeightSSIM}{0.852}
\newcommand{\valBaseLargeHRRCeightLPIPS}{0.359}
\newcommand{\valCoreLargeHRRCeightPSNR}{24.266}
\newcommand{\valCoreLargeHRRCeightSSIM}{0.867}
\newcommand{\valCoreLargeHRRCeightLPIPS}{0.336}

\newcommand{\valMvDLPSNR}{24.545}
\newcommand{\valMvDLSSIM}{0.787}
\newcommand{\valMvDLLPIPS}{0.149}
\newcommand{\valMvDLtwoPSNR}{18.745}
\newcommand{\valMvDLtwoSSIM}{0.710}
\newcommand{\valMvDLtwoLPIPS}{0.296}
\newcommand{\valMvDLfourPSNR}{16.322}
\newcommand{\valMvDLfourSSIM}{0.644}
\newcommand{\valMvDLfourLPIPS}{0.454}
\newcommand{\valMvDLeightPSNR}{15.639}
\newcommand{\valMvDLeightSSIM}{0.672}
\newcommand{\valMvDLeightLPIPS}{0.497}
\newcommand{\valTransDLPSNR}{24.711}
\newcommand{\valTransDLSSIM}{0.792}
\newcommand{\valTransDLLPIPS}{0.152}
\newcommand{\valTransDLtwoPSNR}{20.081}
\newcommand{\valTransDLtwoSSIM}{0.743}
\newcommand{\valTransDLtwoLPIPS}{0.250}
\newcommand{\valTransDLfourPSNR}{18.052}
\newcommand{\valTransDLfourSSIM}{0.722}
\newcommand{\valTransDLfourLPIPS}{0.371}
\newcommand{\valTransDLeightPSNR}{17.485}
\newcommand{\valTransDLeightSSIM}{0.768}
\newcommand{\valTransDLeightLPIPS}{0.419}
\newcommand{\valDepthDLPSNR}{28.145}
\newcommand{\valDepthDLSSIM}{0.905}
\newcommand{\valDepthDLLPIPS}{0.083}
\newcommand{\valDepthDLtwoPSNR}{16.780}
\newcommand{\valDepthDLtwoSSIM}{0.654}
\newcommand{\valDepthDLtwoLPIPS}{0.410}
\newcommand{\valDepthDLfourPSNR}{12.747}
\newcommand{\valDepthDLfourSSIM}{0.400}
\newcommand{\valDepthDLfourLPIPS}{0.591}
\newcommand{\valDepthDLeightPSNR}{11.529}
\newcommand{\valDepthDLeightSSIM}{0.368}
\newcommand{\valDepthDLeightLPIPS}{0.631}
\newcommand{\valHiDLPSNR}{26.497}
\newcommand{\valHiDLSSIM}{0.871}
\newcommand{\valHiDLLPIPS}{0.109}
\newcommand{\valHiDLtwoPSNR}{24.072}
\newcommand{\valHiDLtwoSSIM}{0.836}
\newcommand{\valHiDLtwoLPIPS}{0.189}
\newcommand{\valHiDLfourPSNR}{22.900}
\newcommand{\valHiDLfourSSIM}{0.834}
\newcommand{\valHiDLfourLPIPS}{0.263}
\newcommand{\valHiDLeightPSNR}{22.578}
\newcommand{\valHiDLeightSSIM}{0.878}
\newcommand{\valHiDLeightLPIPS}{0.303}
\newcommand{\valPixDLPSNR}{27.203}
\newcommand{\valPixDLSSIM}{0.882}
\newcommand{\valPixDLLPIPS}{0.103}
\newcommand{\valPixDLtwoPSNR}{25.757}
\newcommand{\valPixDLtwoSSIM}{0.847}
\newcommand{\valPixDLtwoLPIPS}{0.192}
\newcommand{\valPixDLfourPSNR}{24.342}
\newcommand{\valPixDLfourSSIM}{0.832}
\newcommand{\valPixDLfourLPIPS}{0.307}
\newcommand{\valPixDLeightPSNR}{23.785}
\newcommand{\valPixDLeightSSIM}{0.866}
\newcommand{\valPixDLeightLPIPS}{0.350}
\newcommand{\valBaseDLPSNR}{27.421}
\newcommand{\valBaseDLSSIM}{0.891}
\newcommand{\valBaseDLLPIPS}{0.105}
\newcommand{\valBaseDLtwoPSNR}{25.495}
\newcommand{\valBaseDLtwoSSIM}{0.834}
\newcommand{\valBaseDLtwoLPIPS}{0.265}
\newcommand{\valBaseDLfourPSNR}{18.721}
\newcommand{\valBaseDLfourSSIM}{0.587}
\newcommand{\valBaseDLfourLPIPS}{0.576}
\newcommand{\valBaseDLeightPSNR}{14.388}
\newcommand{\valBaseDLeightSSIM}{0.456}
\newcommand{\valBaseDLeightLPIPS}{0.669}
\newcommand{\valCoreDLPSNR}{28.389}
\newcommand{\valCoreDLSSIM}{0.903}
\newcommand{\valCoreDLLPIPS}{0.082}
\newcommand{\valCoreDLtwoPSNR}{26.714}
\newcommand{\valCoreDLtwoSSIM}{0.860}
\newcommand{\valCoreDLtwoLPIPS}{0.173}
\newcommand{\valCoreDLfourPSNR}{23.129}
\newcommand{\valCoreDLfourSSIM}{0.770}
\newcommand{\valCoreDLfourLPIPS}{0.354}
\newcommand{\valCoreDLeightPSNR}{20.671}
\newcommand{\valCoreDLeightSSIM}{0.765}
\newcommand{\valCoreDLeightLPIPS}{0.449}

\newcommand{\anisoDoneLargeLate}{0.580}
\newcommand{\blackDoneLargeEight}{1.11}
\newcommand{\valDoneLargePSNR}{27.544}

\newcommand{\valDoneLargeHRRCeightPSNR}{23.850}

\newcommand{\blackCoreLargeEight}{0.78}

\newcommand{\blackBaseLargeEight}{1.00}

\title{PaveSplat: Resolution-Robust Feed-Forward Gaussian Splatting with
  Residually Corrected Surface-Aligned Covariances}
\titlerunning{PaveSplat}

\author{Janghun Kim\textsuperscript{({\large \Letter})}\orcidlink{0000-0003-2087-2191} \and Hyun-Cheol Kim\orcidlink{0000-0001-6843-5949} \and Eunbin An\orcidlink{0000-0002-8514-2432} \and Myeonghun Oh\orcidlink{0009-0001-9935-5922} \and
  Hyon-Gon Choo\orcidlink{0000-0002-0742-5429}}
\authorrunning{J. Kim et al.}
\institute{ETRI, South Korea\\
\email{\{janghun, kimhc, eunbin.an, mhun9801, hyongonchoo\}@etri.re.kr}}

\begin{document}
\maketitle

\begin{abstract}
We present \emph{PaveSplat}, a feed-forward Gaussian splatting method that adds
a learned residual to the analytic surface-aligned covariance, keeping surfels
paving the visible surface across rendering resolutions. Pixel-aligned feed-forward Gaussian
splatting predicts one primitive per pixel from a few images in a single
pass, and its rendering quality depends on how well these primitives cover the
surface. Recent surfel-based methods compute the surface-aligned covariance
analytically from predicted depth. However, this covariance has no way to correct
the surfel orientation where the depth-derived frame is ill-posed. The surfels
then cover the surface loosely; at the training resolution each surfel's one-pixel
minimum footprint covers the gaps, but they show up as holes when the same prediction is
re-rendered at a higher resolution. To address this, we keep the analytic
covariance as an initialization, with its axes aligned to the surface tangents,
and add an identity-initialized unit-quaternion residual that reorients a surfel
only where this improves rendering, turning surfels that the analytic frame
mis-aims back onto the surface. On RealEstate10K, PaveSplat achieves the best PSNR, SSIM, and LPIPS
among single-Gaussian-per-pixel methods at every scale ($1\times$--$8\times$).
The native-resolution gain is small; the benefit appears when the same
prediction is rendered at higher display resolutions without re-running the network.
On zero-shot DL3DV, it leads PSNR within this group at every scale and ranks
first or second on SSIM/LPIPS. A coverage-hole analysis shows fewer uncovered pixels than
both baselines under high-resolution re-rendering.
\end{abstract}

\section{Introduction}
\label{sec:intro}

\begin{figure}[t]
  \centering
  \includegraphics[width=0.95\linewidth]{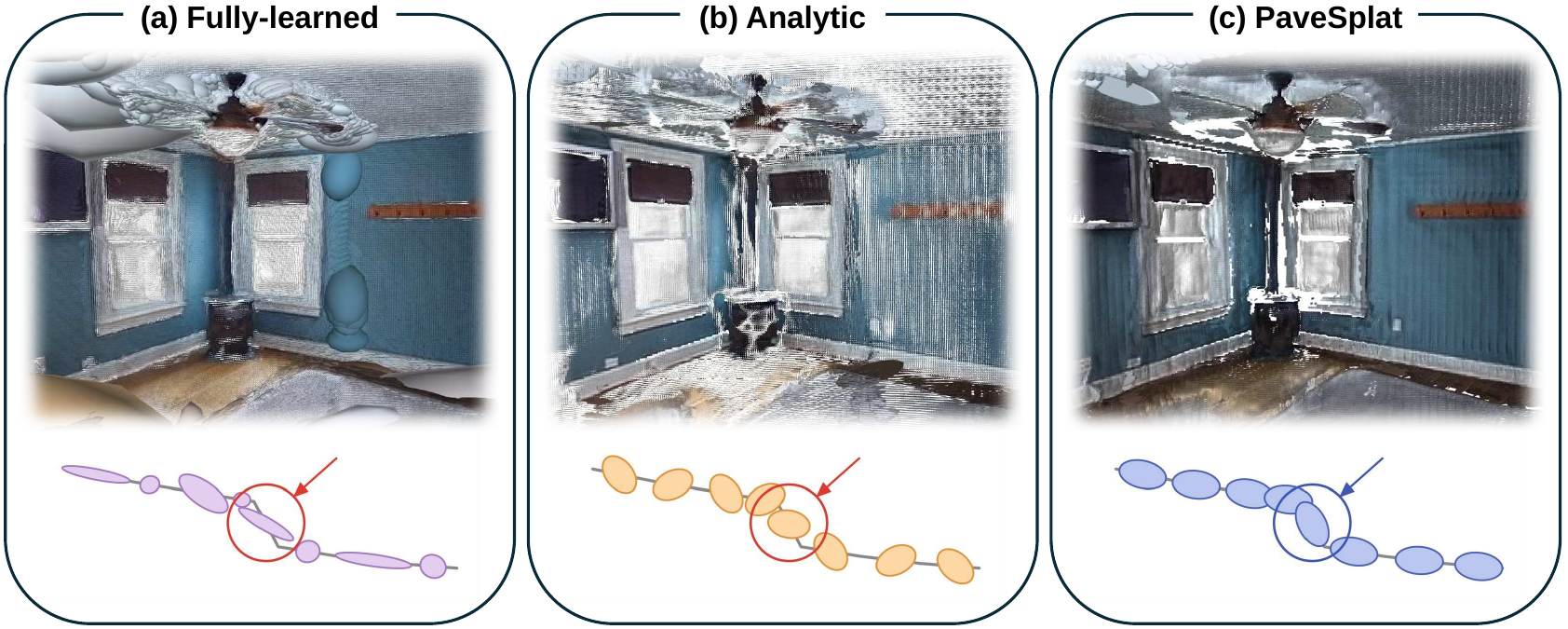}
  \caption{\textbf{Three ways to cover a surface with surfels.} Per regime: the
  per-pixel Gaussians for one scene as covariance ellipsoids from a novel view
  (top) and a schematic of how surfels lie on the surface (bottom).
  \textbf{(a)}~Fully-learned covariances (DepthSplat~\cite{xu2025depthsplat}) leave
  scale uncontrolled, so oversized ellipsoids over-spread while small ones fail to
  tile. \textbf{(b)}~The analytic prior (SurfSplat~\cite{he2026surfsplat}) sets the
  frame from depth with no pathway to correct it, leaving anisotropic surfels
  mis-oriented to the tangents. \textbf{(c)}~PaveSplat makes the analytic frame
  per-pixel correctable ($\qfin{=}\qana\otimes\dq$), so surfels reorient to pave the
  surface. This isolates surfel shape and coverage.}
  \label{fig:teaser}
\end{figure}

Feed-forward Gaussian splatting reconstructs a renderable 3D scene from a few
images in a single network pass. Pixel-aligned primitives are predicted
from image features~\cite{charatan2024pixelsplat,chen2024mvsplat,xu2025depthsplat}
and rendered with a differentiable rasterizer~\cite{kerbl3dgs}. Among these
methods, how each primitive's \emph{covariance} is obtained spans two regimes.
First, fully learned covariances --- a free quaternion and scale
regressed per pixel~\cite{charatan2024pixelsplat,chen2024mvsplat} --- provide
high flexibility but no explicit surface grounding. With nothing tying the scale
to the local surface, the per-pixel sizes vary without control: some surfels grow
into long needles that over-spread into blur, while others stay too small to fill
the surface (Fig.~\ref{fig:teaser}(a)).
This looks adequate at the training resolution, but the uneven, surface-agnostic scales
leave holes once the prediction is re-rendered at a higher resolution. Second, the
analytic regime, SurfSplat~\cite{he2026surfsplat} derives the covariance
\emph{analytically} from predicted depth: surface tangents come from the
unprojected point map, their cross product gives a normal, and the released
SurfSplat normal-alignment step (described in that paper as a Rodrigues rotation)
ties the surfel frame to that local surface. This imposes a surface-continuity
prior that makes the covariance tied to the predicted local surface.

The analytic regime, however, trades \emph{correctability} for its prior: the covariance is a deterministic function of predicted
depth. Where the depth-derived frame is ill-posed --- at depth edges, thin
structures, or textureless slopes --- the surfel orientation can be misaligned;
since the only trainable covariance parameters are two bounded scale multipliers,
there is no learned orientation variable at a given predicted depth. The prior also
leaves an in-plane misalignment: the analytic rotation is built from the normal
alone, so its two in-plane axes are fixed by the normal rather than aligned with
the tangent directions from which the radii are measured. This would be
harmless for an isotropic disk, but the
in-plane radii are anisotropic: an elongated surfel has a meaningful long axis,
and nothing ties it to the surface tangents it was derived from. The
``surface-aligned'' covariance is therefore not aligned with those tangents
(Fig.~\ref{fig:teaser}(b)). On a
surface seen at an oblique angle the two tangents differ in length, so a surfel
whose long axis falls across the short tangent over-spreads there while leaving
gaps along the long one. Such surfels cover the surface loosely --- adequate at
the training resolution but exposed as holes once the prediction is re-rendered
at a higher resolution.

This paper presents \textbf{PaveSplat}, a \emph{residually corrected
surface-aligned covariance} that places anisotropic surfels between the fully
learned and fully analytic regimes, paving the visible surface
(Fig.~\ref{fig:teaser}). Unlike the
analytic baseline, our approach does not treat the analytic prior as a hard
constraint; we keep it as an \emph{initialization} and act on the two separable
aspects of a surfel's covariance, its in-plane extent and its orientation. For
\emph{shape}, we keep the analytic in-plane radii and their bounded multipliers,
but build the frame so its in-plane axes align with the projected surface
tangents; we refer to this geometry-consistent stage as the \emph{surface-aligned
prior}. For \emph{orientation}, we add a \emph{residual reorientation}, our only
newly learned covariance parameter: a per-pixel rotation residual $\dq$ that
post-composes the analytic rotation and is initialized to identity. The first
forward pass thus reproduces the analytic frame exactly, and training reorients a
surfel only where rendering evidence demands (Fig.~\ref{fig:teaser}(c)),
preserving the anisotropic surface support that degenerate surfels lose, so that
coverage holds when the prediction is re-rendered at a higher resolution.

Consequently, PaveSplat does not replace the surface prior with a free
covariance; it opens only the missing orientation degree of freedom around an
analytic operating point. We therefore measure whether the residual is used or
collapses to identity: we log the residual rotation-angle distribution throughout
training, report full-test per-scene statistics against a sample-matched SurfSplat
under a single controlled recipe (the covariance derivation is the only variable),
and instrument the geometry pipeline with diagnostics that tie the metric
differences to the mechanism.

Our contributions are summarized as follows:
\begin{itemize}
  \item a \emph{residually corrected surface-aligned covariance} for feed-forward
  surfel splatting that keeps the analytic extent prior and corrects only the
  surfel \emph{orientation} with a lightweight, identity-initialized quaternion
  residual, placed between the fully learned and fully analytic regimes;
  \item a \emph{geometry-consistent analytic initialization} of that prior --- a
  tangent-aligned frame with explicit degeneracy handling --- so the in-plane
  anisotropy acts along the surface tangents it was measured on rather than along
  axes fixed by the normal alone;
  \item \emph{mechanistic validation} beyond image metrics: the residual does not
  collapse, in-plane anisotropy and coverage holes move as predicted, and PaveSplat
  reports the best RealEstate10K PSNR, SSIM, and LPIPS among single-Gaussian-per-pixel
  methods at every render scale ($1\times$--$8\times$), winning on most test scenes against
  the same-recipe analytic baseline at each scale;
  on zero-shot ScanNet, a ground-truth normal check (supplement) further shows
  lower rendered-normal error, consistent with reorienting covariance for coverage.
\end{itemize}

\section{Related work}
\label{sec:related}

\subsection{Gaussian splatting and surface-aligned primitives}
3D Gaussian splatting~\cite{kerbl3dgs} represents a scene with anisotropic
Gaussian primitives and renders them with an EWA-style differentiable
rasterizer~\cite{zwicker2001ewa}. However, per-scene optimization and single-pass prediction constrain each
primitive's covariance very differently. Surface-oriented variants add geometric structure to the per-scene
setting: 2DGS~\cite{huang2dgs} flattens Gaussians into disks; Gaussian
surfels~\cite{dai2024surfels} use surfel primitives for surface
reconstruction; SuGaR~\cite{guedon2024sugar} regularizes Gaussians toward an
extractable surface. These methods optimize each scene from many posed views, so
the primitive support is adjusted through repeated rendering losses. In
feed-forward reconstruction, the same kind of support must be predicted from a
few posed images without per-scene refinement.

\subsection{Feed-forward sparse-view novel view synthesis}
Generalizable novel view synthesis predicts a renderable scene from sparse posed
images, with early systems built on NeRF-style radiance
fields~\cite{mildenhall2020nerf,yu2021pixelnerf,chen2021mvsnerf,suhail2022gpnr,du2023attnrend,xu2024murf}.
However, replacing iterative per-scene fitting with a feed-forward predictor
makes the primitive covariance a learned quantity inferred from limited
evidence. pixelSplat~\cite{charatan2024pixelsplat} introduced pixel-aligned
Gaussian primitives to this regime; MVSplat~\cite{chen2024mvsplat} uses
plane-sweep cost volumes~\cite{yao2018mvsnet,wang2026mvssurvey}; TranSplat~\cite{zhang2025transplat}
and HiSplat~\cite{tang2025hisplat} refine matching and hierarchy;
DepthSplat~\cite{xu2025depthsplat} connects the Gaussian head to monocular and
multi-view depth cues~\cite{yang2024depthanythingv2,xu2023unimatch}; Splatter
Image~\cite{szymanowicz2024splatter}, GS-LRM~\cite{zhang2024gslrm}, and
SparseSplat~\cite{zhang2026sparsesplat} change the prediction layout. In these
Gaussian feed-forward methods, rotation and scale are commonly regressed as free
per-pixel outputs. This gives the network capacity, but the covariance is not
explicitly tied to the local surface frame.

\subsection{Analytic vs.\ learned covariance}
A Gaussian covariance defines the spatial support used to cover the predicted
surface, and therefore forms a separate design axis from depth, opacity, and
color. SurfSplat~\cite{he2026surfsplat} moves from free covariance regression to
a feed-forward surfel representation: the local tangents and normal are computed
from predicted depth, the surfel rotation and in-plane scales are derived
analytically, and alpha blending is constrained to use multiple layers. However,
this analytic choice also makes the covariance a deterministic function of the
depth-derived frame. Where that frame is ambiguous, the representation has no
learned orientation parameter analogous to the free quaternion used by
pixel-aligned 3DGS predictors. Residual refinement has been used in feed-forward
pipelines for positions or features~\cite{tang2025hisplat}, but covariance
orientation remains less directly separated from the analytic surface prior.

While existing feed-forward Gaussian methods either learn the covariance freely
or derive it analytically from predicted depth, PaveSplat treats the analytic
surfel covariance as an operating point. SurfSplat is the same-recipe analytic
baseline in our experiments; we align the in-plane axes with the measured
tangents and learn only an identity-initialized residual orientation around that
frame (Sec.~\ref{sec:method}).

\section{Method}
\label{sec:method}

\begin{figure}[t]
  \centering
  \includegraphics[width=0.95\linewidth]{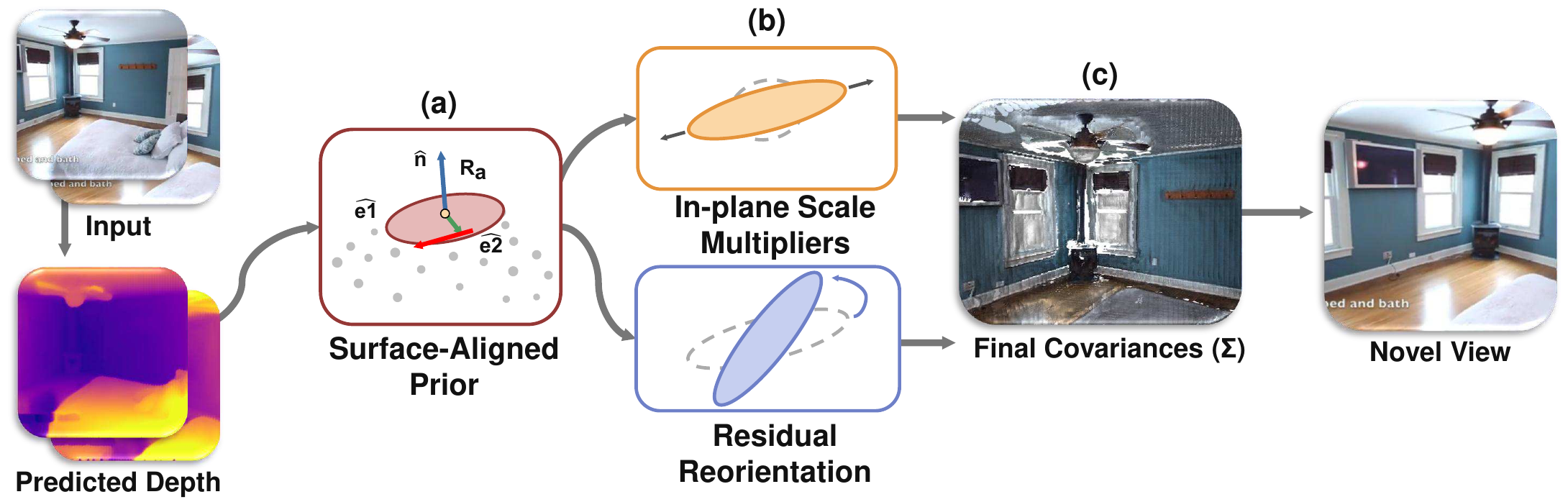}
  \caption{\textbf{Overview of PaveSplat covariance construction.} From the predicted
  depth, \textbf{(a)}~the \emph{surface-aligned prior} sets the analytic frame
  $\Rana{=}[\eone\,\etwo\,\nhat]$, supplying tangent-aligned shape and the initial
  orientation. \textbf{(b)}~Two per-primitive refinements act on this prior: in-plane
  scale multipliers (orange) set the anisotropic radii $s_i{=}\bar s_i\,m_i$, and a
  residual reorientation (blue) --- the only added learned covariance parameter ---
  adapts orientation via $\qfin{=}\qana{\otimes}\dq$. \textbf{(c)}~The two recombine
  as $\Rfin{=}\Rana\Rres$ into the final covariance $\Sfin$, rendered to the novel
  view.}
  \label{fig:overview}
\end{figure}

\subsection{Preliminaries}
\label{sec:method-prelim}

We first briefly review the feed-forward surfel formulation we build on. Given
$V$ input images $\mathbf{I}^v$ with camera intrinsics $\mathbf{K}^v$ and
camera-to-world extrinsics $\mathbf{T}^v$, a feed-forward Gaussian-splatting
network predicts one pixel-aligned primitive per pixel,
\begin{equation}
  f_\theta:\ \{(\mathbf{I}^v,\mathbf{K}^v,\mathbf{T}^v)\}_{v=1}^{V}
  \;\longmapsto\;
  \bigl\{(\boldsymbol{\mu}_j,\alpha_j,\qfin_j,\mathbf{s}_j,\mathbf{c}_j)\bigr\}_{j=1}^{V\!\times\!H\!\times\!W},
  \label{eq:ff-map}
\end{equation}
with center $\boldsymbol{\mu}$, opacity $\alpha$, rotation $\qfin$, per-axis
scales $\mathbf{s}$, and spherical-harmonics color $\mathbf{c}$, rendered by a
differentiable rasterizer~\cite{kerbl3dgs}. We follow the surfel (2D Gaussian)
formulation~\cite{huang2dgs,he2026surfsplat}: the third scale is pinned to a
small constant, so each primitive is a thin disk whose orientation
\emph{is} the local surface geometry. Depth comes from the DepthSplat
backbone~\cite{xu2025depthsplat} (a plane-sweep cost
volume~\cite{yao2018mvsnet,chen2024mvsplat} fused with monocular
features~\cite{yang2024depthanythingv2}), which we keep unchanged; the same
per-pixel head also emits the opacity, the two in-plane scale multipliers, the
color, and the rotation parameters from which the covariance is built. The center is
$\boldsymbol{\mu}(\pix)=\mathbf{o}(\pix)+z(\pix)\,\rdir(\pix)$, where
$\rdir$ is the $z$-normalized world ray of pixel $\pix$ and $z$ its predicted
depth.

We keep the rest of the surfel pipeline fixed to
SurfSplat~\cite{he2026surfsplat}, so that the covariance derivation is the only
independent variable: the $0.6$ opacity cap that forces multi-layer alpha
compositing instead of a few near-opaque primitives, the alpha-normalized
differentiable surfel rasterizer~\cite{kerbl3dgs,huang2dgs}, and the photometric
loss $\mathcal{L}=\mathrm{MSE}+0.05\,\mathrm{LPIPS}$~\cite{zhang2018lpips} on
rendered target views are all unchanged (alpha thresholds in the supplement).
Only the derivation of rotation, scale, and covariance from the predicted depth
differs; the rest of this section describes it.

\paragraph{Analytic surface covariance and its missing degree of freedom.}
SurfSplat~\cite{he2026surfsplat} derives the surfel rotation analytically: it
estimates surface tangents from the depth-unprojected point map, takes their
cross product as the normal $\nhat$, and applies the released SurfSplat
canonical-to-normal alignment construction, described in SurfSplat as Rodrigues'
formula, imposing a surface-continuity prior.
The resulting covariance, however, is a \emph{deterministic function of
predicted depth}. The depth branch is fine-tuned through the rendering loss, so
the image-to-covariance mapping as a whole is still learned. The restriction lies
in the depth-to-covariance parameterization: at a fixed predicted depth it has no
learned orientation variable, since the only learned covariance quantities are
two bounded in-plane scale multipliers. One limit remains however accurate the
depth is: a finite-difference (Sobel) tangent can fail across depth
discontinuities, where the differenced pixels lie on no common local surface, and
at grazing incidence, where foreshortening destabilizes the estimate. In addition,
a normal-only rotation fixes the in-plane axes by the normal alone, so they are
not aligned with the tangent directions from which the two anisotropic in-plane
radii are measured. This rigidity can leave surfels round
and near-isotropic; their incomplete surface support becomes clearest when the
prediction is re-rendered at higher resolution.

PaveSplat keeps the analytic covariance as an operating point rather than a hard
constraint, but makes it
\emph{(i)} geometrically consistent and \emph{(ii)} adjustable by a lightweight
learned rotation residual, organized around the two aspects of a surfel's
covariance: a surface-aligned prior (Sec.~\ref{sec:method-frame}) sets the
in-plane shape, and a residual reorientation (Sec.~\ref{sec:method-residual})
adapts orientation. As Fig.~\ref{fig:overview}(a) shows, the single analytic frame
$\Rana=[\eone\,\etwo\,\nhat]$ supplies both rotation and anisotropic radii
$\bar s_1,\bar s_2$.

\subsection{Surface-aligned prior: geometry-consistent initialization}
\label{sec:method-frame}

\paragraph{Camera-space formulation.}
We construct the surface frame in camera coordinates and transform the
resulting covariance to world coordinates exactly once. For pixel
$\pix=(x,y)$ with homogeneous coordinate $\tilde{\pix}$, the camera-space
point map is
\begin{equation}
  \ptmap(\pix) \;=\; z(\pix)\,
  \frac{\mathbf{K}^{-1}\tilde{\pix}}{\bigl\langle \mathbf{K}^{-1}\tilde{\pix} \bigr\rangle_z},
  \label{eq:pointmap}
\end{equation}
where $\langle\cdot\rangle_z$ denotes the $z$-component. In this frame the
optical axis is $+z$, so the fixed camera-space hemisphere convention used below
--- a normal with $\langle\nhat\rangle_z\ge 0$ --- is well-defined. This
coordinate-consistent formulation keeps the rotation semantics of the analytic
frame intact. We treat it as a well-posed construction of the analytic frame
rather than an independent experimental variable: it is applied identically in
all of our configurations, while our same-recipe SurfSplat baseline reproduces
the released construction.

\paragraph{Tangent-aligned frame.}
Surface tangents $\tu=\partial\ptmap/\partial x$ and
$\tv=\partial\ptmap/\partial y$ are estimated with $3{\times}3$ Sobel
derivative filters on the point map, as in SurfSplat. From these we build a
proper orthonormal frame whose in-plane axes follow the tangents:
\begin{equation}
  \nhat=\frac{\tu\times\tv}{\lVert\tu\times\tv\rVert},\qquad
  \eone=\frac{\tu-(\tu\!\cdot\!\nhat)\,\nhat}{\lVert\tu-(\tu\!\cdot\!\nhat)\,\nhat\rVert},\qquad
  \etwo=\nhat\times\eone,
  \label{eq:frame}
\end{equation}
where we flip $\nhat$ into the hemisphere $\langle\nhat\rangle_z\ge 0$ before
forming $\etwo=\nhat\times\eone$, so $\Rana=[\eone\;\etwo\;\nhat]$ stays
right-handed ($\det\Rana=+1$). The flip only fixes this hemisphere; it does not
enforce neighborhood, cross-view, or global orientation consistency. Two degeneracies ---
near-colinear tangents and a tangent near-parallel to the normal --- are handled
with explicit fallbacks (supplement). The analytic in-plane radii are the
tangent magnitudes on these axes,
\begin{equation}
  \sbarone=\lVert\tu\rVert,\qquad
  \sbartwo=\bigl\lVert \tv-(\tv\!\cdot\!\eone)\,\eone \bigr\rVert,
  \label{eq:radii}
\end{equation}
Before these norms, each tangent component is clamped to $[-0.1,0.1]$, matching
the baseline guard against large derivatives at depth edges. As in SurfSplat,
two bounded learned multipliers refine them, $s_i=\bar{s}_i\, m_i$ with
$m_i=\mathrm{clamp}\bigl(\mathrm{softplus}(\cdot),[\tfrac13,3]\bigr)$, and the
thin axis $s_{\epsilon}$ is pinned to the global scale floor. The frame in
Eq.~\ref{eq:frame} keeps the normal of a normal-only frame built from the same
point map --- only the in-plane axes change --- so the anisotropic radii now act
along the directions they were measured on.

\subsection{Residual reorientation: the learned covariance correction}
\label{sec:method-residual}

Section~\ref{sec:method-frame} sets a surfel's in-plane \emph{shape}: the
radii remain anisotropic through the two bounded scale multipliers $m_i$
(following SurfSplat~\cite{he2026surfsplat}) and now act along the
tangent-aligned axes. The analytic prior still lacks a learned pathway for the
other aspect, its \emph{orientation}: the frame $\Rana$ depends on the
depth-derived tangents, so where that frame is ill-posed an otherwise
well-shaped surfel can be oriented away from the surface and provide incomplete
support. To address this, we add a learned orientation pathway --- our only new learned
covariance parameter --- as a lightweight \emph{rotation residual}
(Fig.~\ref{fig:overview}(b)): a per-pixel
unit quaternion $\dq$ composed \emph{on the right} of the analytic frame:
\begin{equation}
  \Rfin \;=\; \Rana\, \Rres(\dq),
  \qquad \text{equivalently}\qquad
  \qfin = \qana \otimes \dq ,
  \label{eq:residual}
\end{equation}
where $\Rres(\dq)$ is the rotation matrix of $\dq$ and $\otimes$ is the
Hamilton product. Right-composition expresses $\dq$ \emph{in the local analytic
frame}: its axis-angle content is interpretable relative to the surface
($\eone,\etwo,\nhat$), not to an arbitrary global frame.

The reoriented normal $\nhat_c=\mathrm{normalize}(\Rfin\,\mathbf{e}_3)$ --- the
analytic $\nhat$ of Eq.~\ref{eq:frame} reoriented by the residual --- is stored
as the surfel normal and sets the covariance orientation used by the renderer.
Thus the residual receives gradients through the differentiable rendering loss
wherever changing covariance orientation changes the rendered image.
The camera-space covariance and its world transform (Fig.~\ref{fig:overview}(c)) are
\begin{equation}
  \Scov=\Rfin\,\mathrm{diag}(s_1,s_2,s_{\epsilon})^2\,\Rfin^{\!\top},
  \qquad
  \Sfin^{\mathrm{w}}=\cam\,\Scov\,\cam^{\!\top},
  \label{eq:cov}
\end{equation}
where $\cam$ is the rotation of the camera-to-world extrinsics $\mathbf{T}^v$,
applied exactly once; the supplement gives the full per-pixel construction as
pseudocode, from the camera-space point map to this single world transform.

\paragraph{Prediction head.}
The residual reuses the per-pixel Gaussian head unchanged: alongside the existing
opacity, scale-multiplier, and color outputs, it emits four additional channels
normalized to the unit quaternion $\dq$, adding only $9{,}512$ parameters
($<0.05\%$ of the model). These
channels are \emph{identity-initialized} (zero weights, identity-quaternion
bias), so $\dq{=}\mathrm{id}$ on the first forward pass and the network reproduces
the analytic frame exactly before training. Training then moves the rotation only
where the photometric evidence asks for it, starting from the analytic operating
point rather than a random rotation. This keeps the correction deliberately
low-capacity: the residual can rotate the local covariance, but it leaves the
predicted depth, center, opacity, rasterizer, and loss unchanged. We therefore
interpret it as a robustness parameterization around SurfSplat's surface prior,
not as a learned replacement for the surface estimate or a parameter-reduction
mechanism.

\paragraph{Why a residual, not a free rotation.}
A fully learned rotation (as in pixel-aligned 3DGS
methods~\cite{charatan2024pixelsplat,chen2024mvsplat}) discards the surface
prior entirely; a fixed analytic rotation keeps the prior but cannot adapt its
orientation. In contrast, the identity-initialized residual interpolates between
the two regimes with the analytic frame as the operating point, and keeps the
normal axis
\emph{identifiable}: $\nhat_c$ remains an explicit, differentiable orientation
variable in the rendering graph rather than an emergent property of an
unconstrained quaternion. Whether the residual
moves away from identity is an empirical question we measure directly
(Sec.~\ref{sec:experiments}); a collapsed residual ($\dq\!\approx\!\mathrm{id}$)
would reduce our method to its analytic initialization.

\section{Experiments}
\label{sec:experiments}

\subsection{Experimental setup}
\label{sec:exp-setup}

\paragraph{Datasets and metrics.}
We train and evaluate on RealEstate10K~\cite{zhou2018stereo} at
$256{\times}256$ with two input views, under the standard evaluation protocol
of pixelSplat~\cite{charatan2024pixelsplat} (test split and target-view index
shared by~\cite{chen2024mvsplat,zhang2025transplat,tang2025hisplat,he2026surfsplat}).
We report PSNR, SSIM~\cite{wang2004ssim}, and LPIPS~\cite{zhang2018lpips}.

\paragraph{Implementation.}
All rows we train use the official DepthSplat RE10K recipe~\cite{xu2025depthsplat}
with a Depth Anything V2 ViT-L~\cite{yang2024depthanythingv2} backbone and the
inherited loss and rendering pipeline of Sec.~\ref{sec:method-prelim}. Optimizer,
schedule, head settings, and the ViT-S/B/L ladder are in the supplement. The
controlled SurfSplat row uses the same total sample count~\cite{he2026surfsplat},
so it is sample-matched to the published model.

\paragraph{Controlled comparisons.}
Every row we train shares the inherited pipeline of Sec.~\ref{sec:method-prelim},
with byte-identical data pipeline and schedule. The independent variable is how
rotation, scale, and covariance are derived.

\paragraph{Statistical protocol.}
We report the controlled comparison \emph{per scene}: at the final checkpoint
we score both models over the full test set and report the fraction of
scenes PaveSplat wins and a two-sided Wilcoxon signed-rank test,
Holm-corrected within each metric family. Every per-scene image-metric
comparison we report below reaches $p{<}10^{-10}$ under this test; we therefore
cite the win fractions in the text and do not restate the test at each result.

\paragraph{Reporting conventions.}
Unless noted, main-paper results use the ViT-L backbone and two input views, and
report the $256^2$ \emph{Standard} scale together with $512^2/1024^2/2048^2$
high-resolution re-renders ($2\times/4\times/8\times$, HRRC), a diagnostic
stress test where the same prediction is rasterized on larger grids and scored
against bicubically-upsampled targets~\cite{he2026surfsplat}. SurfSplat$^{\ddagger}$
is our same-recipe reproduction (the analytic baseline), not the released
checkpoint. In tables, \textbf{bold} marks the column-best among
single-Gaussian-per-pixel methods and \underline{underline} (where used) the
second-best; gray cells, where present, are quoted from the cited paper.

\subsection{Novel-view synthesis at native resolution}
\label{sec:exp-main}

\begin{table}[t]
  \centering
  \caption{\textbf{Novel view synthesis on RealEstate10K across render
  resolutions.} All cells are our measurements on the common index;
  multi-Gaussian methods (pixelSplat, HiSplat) are ruled off above the
  single-Gaussian-per-pixel group. SurfSplat$^\ddagger$: our same-recipe reproduction.}
  \label{tab:main}
  \setlength{\tabcolsep}{3.2pt}
  \resizebox{\textwidth}{!}{%
  \begin{tabular}{@{}lccccccccccccc@{}}
    \toprule
    & \multicolumn{3}{c}{$256^2$ (\textbf{Standard})} & \multicolumn{3}{c}{$512^2$ (HRRC)}
    & \multicolumn{3}{c}{$1024^2$ (HRRC)} & \multicolumn{3}{c}{$2048^2$ (HRRC)} \\
    \cmidrule(lr){2-4}\cmidrule(lr){5-7}\cmidrule(lr){8-10}\cmidrule(lr){11-13}
    Method & PSNR$\uparrow$ & SSIM$\uparrow$ & LPIPS$\downarrow$
           & PSNR$\uparrow$ & SSIM$\uparrow$ & LPIPS$\downarrow$
           & PSNR$\uparrow$ & SSIM$\uparrow$ & LPIPS$\downarrow$
           & PSNR$\uparrow$ & SSIM$\uparrow$ & LPIPS$\downarrow$ \\
    \midrule
    pixelSplat~\cite{charatan2024pixelsplat} & \valPixPSNR & \valPixSSIM & \valPixLPIPS & \valPixTwoPSNR & \valPixTwoSSIM & \valPixTwoLPIPS & \valPixFourPSNR & \valPixFourSSIM & \valPixFourLPIPS & \valPixEightPSNR & \valPixEightSSIM & \valPixEightLPIPS \\
    HiSplat~\cite{tang2025hisplat}           & \valHiPSNR & \valHiSSIM & \valHiLPIPS & \valHiTwoPSNR & \valHiTwoSSIM & \valHiTwoLPIPS & \valHiFourPSNR & \valHiFourSSIM & \valHiFourLPIPS & \valHiEightPSNR & \valHiEightSSIM & \valHiEightLPIPS \\
    \midrule
    MVSplat~\cite{chen2024mvsplat}           & \valMvPSNR & \valMvSSIM & \valMvLPIPS & \valMvTwoPSNR & \valMvTwoSSIM & \valMvTwoLPIPS & \valMvFourPSNR & \valMvFourSSIM & \valMvFourLPIPS & \valMvEightPSNR & \valMvEightSSIM & \valMvEightLPIPS \\
    TranSplat~\cite{zhang2025transplat}      & \valTransPSNR & \valTransSSIM & \valTransLPIPS & \valTransTwoPSNR & \valTransTwoSSIM & \valTransTwoLPIPS & \valTransFourPSNR & \valTransFourSSIM & \valTransFourLPIPS & \valTransEightPSNR & \valTransEightSSIM & \valTransEightLPIPS \\
    DepthSplat~\cite{xu2025depthsplat}       & \valDepthPSNR & \valDepthSSIM & \valDepthLPIPS & \valDepthTwoPSNR & \valDepthTwoSSIM & \valDepthTwoLPIPS & \valDepthFourPSNR & \valDepthFourSSIM & \valDepthFourLPIPS & \valDepthEightPSNR & \valDepthEightSSIM & \valDepthEightLPIPS \\
    SurfSplat$^\ddagger$~\cite{he2026surfsplat} & \valBaseLargePSNR & \valBaseLargeSSIM & \valBaseLargeLPIPS & \valBaseLargeHRRCtwoPSNR & \valBaseLargeHRRCtwoSSIM & \valBaseLargeHRRCtwoLPIPS & \valBaseLargeHRRCfourPSNR & \valBaseLargeHRRCfourSSIM & \valBaseLargeHRRCfourLPIPS & \valBaseLargeHRRCeightPSNR & \valBaseLargeHRRCeightSSIM & \valBaseLargeHRRCeightLPIPS \\
    PaveSplat (ours)                       & \textbf{\valCoreLargePSNR} & \textbf{\valCoreLargeSSIM} & \textbf{\valCoreLargeLPIPS} & \textbf{\valCoreLargeHRRCtwoPSNR} & \textbf{\valCoreLargeHRRCtwoSSIM} & \textbf{\valCoreLargeHRRCtwoLPIPS} & \textbf{\valCoreLargeHRRCfourPSNR} & \textbf{\valCoreLargeHRRCfourSSIM} & \textbf{\valCoreLargeHRRCfourLPIPS} & \textbf{\valCoreLargeHRRCeightPSNR} & \textbf{\valCoreLargeHRRCeightSSIM} & \textbf{\valCoreLargeHRRCeightLPIPS} \\
    \bottomrule
  \end{tabular}%
  }
\end{table}

Table~\ref{tab:main} compares feed-forward methods across the $256^2$ Standard
scale and $512^2$/$1024^2$/$2048^2$ HRRC re-renders. We re-run released
checkpoints under this protocol and the common index; following SurfSplat,
multi-Gaussian-per-pixel methods are ruled off above the
single-Gaussian-per-pixel block.

\afterpage{%
  \begin{figure}[t]
    \centering
    \includegraphics[width=0.95\linewidth]{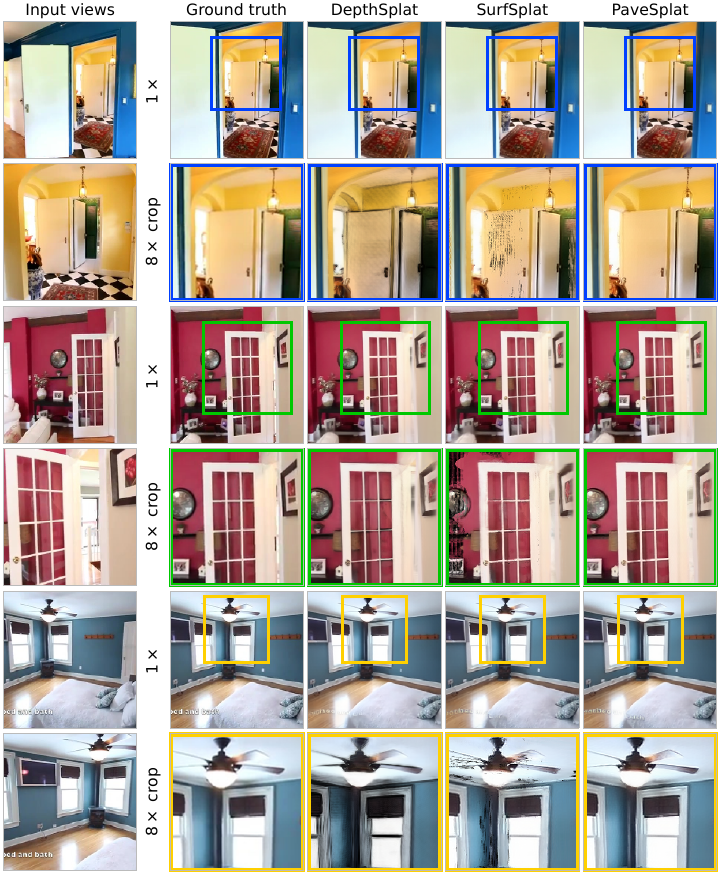}
    \caption{\textbf{Qualitative comparison on RealEstate10K.} Per scene: inputs,
    ground truth, DepthSplat~\cite{xu2025depthsplat}, SurfSplat~\cite{he2026surfsplat},
    and PaveSplat; top row $=$ full-frame native render (crop boxed), bottom row
    $=$ $8\times$ crop. In these
    high-margin cases, PaveSplat keeps more support along thin structures.}
    \label{fig:qualitative}
  \end{figure}%
  \begin{table}[t]
  \centering
  \caption{\textbf{Zero-shot cross-dataset novel view synthesis on DL3DV}~\cite{ling2024dl3dv}.
  RealEstate10K-trained, evaluated zero-shot; all rows are our re-runs.}
  \label{tab:dl3dv}
  \setlength{\tabcolsep}{3.2pt}
  \resizebox{\textwidth}{!}{%
  \begin{tabular}{@{}lccccccccccccc@{}}
    \toprule
    & \multicolumn{3}{c}{Standard ($256{\times}256$)} & \multicolumn{3}{c}{$2\times$ (HRRC)}
    & \multicolumn{3}{c}{$4\times$ (HRRC)} & \multicolumn{3}{c}{$8\times$ (HRRC)} \\
    \cmidrule(lr){2-4}\cmidrule(lr){5-7}\cmidrule(lr){8-10}\cmidrule(lr){11-13}
    Method & PSNR$\uparrow$ & SSIM$\uparrow$ & LPIPS$\downarrow$
           & PSNR$\uparrow$ & SSIM$\uparrow$ & LPIPS$\downarrow$
           & PSNR$\uparrow$ & SSIM$\uparrow$ & LPIPS$\downarrow$
           & PSNR$\uparrow$ & SSIM$\uparrow$ & LPIPS$\downarrow$ \\
    \midrule
    pixelSplat~\cite{charatan2024pixelsplat} & \valPixDLPSNR & \valPixDLSSIM & \valPixDLLPIPS & \valPixDLtwoPSNR & \valPixDLtwoSSIM & \valPixDLtwoLPIPS & \valPixDLfourPSNR & \valPixDLfourSSIM & \valPixDLfourLPIPS & \valPixDLeightPSNR & \valPixDLeightSSIM & \valPixDLeightLPIPS \\
    HiSplat~\cite{tang2025hisplat}           & \valHiDLPSNR & \valHiDLSSIM & \valHiDLLPIPS & \valHiDLtwoPSNR & \valHiDLtwoSSIM & \valHiDLtwoLPIPS & \valHiDLfourPSNR & \valHiDLfourSSIM & \valHiDLfourLPIPS & \valHiDLeightPSNR & \valHiDLeightSSIM & \valHiDLeightLPIPS \\
    \midrule
    MVSplat~\cite{chen2024mvsplat}           & \valMvDLPSNR & \valMvDLSSIM & \valMvDLLPIPS & \valMvDLtwoPSNR & \valMvDLtwoSSIM & \valMvDLtwoLPIPS & \valMvDLfourPSNR & \valMvDLfourSSIM & \valMvDLfourLPIPS & \valMvDLeightPSNR & \valMvDLeightSSIM & \valMvDLeightLPIPS \\
    TranSplat~\cite{zhang2025transplat}      & \valTransDLPSNR & \valTransDLSSIM & \valTransDLLPIPS & \valTransDLtwoPSNR & \valTransDLtwoSSIM & \underline{\valTransDLtwoLPIPS} & \valTransDLfourPSNR & \underline{\valTransDLfourSSIM} & \underline{\valTransDLfourLPIPS} & \underline{\valTransDLeightPSNR} & \textbf{\valTransDLeightSSIM} & \textbf{\valTransDLeightLPIPS} \\
    DepthSplat~\cite{xu2025depthsplat}       & \underline{\valDepthDLPSNR} & \textbf{\valDepthDLSSIM} & \underline{\valDepthDLLPIPS} & \valDepthDLtwoPSNR & \valDepthDLtwoSSIM & \valDepthDLtwoLPIPS & \valDepthDLfourPSNR & \valDepthDLfourSSIM & \valDepthDLfourLPIPS & \valDepthDLeightPSNR & \valDepthDLeightSSIM & \valDepthDLeightLPIPS \\
    SurfSplat$^\ddagger$~\cite{he2026surfsplat} & \valBaseDLPSNR & \valBaseDLSSIM & \valBaseDLLPIPS & \underline{\valBaseDLtwoPSNR} & \underline{\valBaseDLtwoSSIM} & \valBaseDLtwoLPIPS & \underline{\valBaseDLfourPSNR} & \valBaseDLfourSSIM & \valBaseDLfourLPIPS & \valBaseDLeightPSNR & \valBaseDLeightSSIM & \valBaseDLeightLPIPS \\
    PaveSplat (ours) & \textbf{\valCoreDLPSNR} & \underline{\valCoreDLSSIM} & \textbf{\valCoreDLLPIPS} & \textbf{\valCoreDLtwoPSNR} & \textbf{\valCoreDLtwoSSIM} & \textbf{\valCoreDLtwoLPIPS} & \textbf{\valCoreDLfourPSNR} & \textbf{\valCoreDLfourSSIM} & \textbf{\valCoreDLfourLPIPS} & \textbf{\valCoreDLeightPSNR} & \underline{\valCoreDLeightSSIM} & \underline{\valCoreDLeightLPIPS} \\
    \bottomrule
  \end{tabular}%
  }
\end{table}
}

Table~\ref{tab:main} also isolates the comparison central to our claim: the
sample-matched SurfSplat reproduction, where the covariance derivation is the
only independent variable. At the Standard $256^2$ resolution, PaveSplat improves
PSNR from \valBaseLargePSNR{} to \valCoreLargePSNR\,dB, SSIM from
\valBaseLargeSSIM{} to \valCoreLargeSSIM, and LPIPS from \valBaseLargeLPIPS{} to
\valCoreLargeLPIPS. The same ordering holds per scene over the full test set:
PaveSplat is ahead on \psWinPSNRLarge, \psWinSSIMLarge, and
\psWinLPIPSLarge{} of scenes for PSNR, SSIM, and LPIPS, respectively.
This native gain is small, and RE10K image metrics are close to saturation at
this operating point. Our claim concerns display-resolution rendering instead:
the same prediction re-rendered on a denser grid, without re-running the network
(Sec.~\ref{sec:exp-robust}). We therefore read these native metrics together
with the re-rendering, coverage, and ablation results below. The
residual adds \runtimeOverheadPct{} encoder time over the analytic baseline.

This controlled row should be read narrowly. Since the backbone, depth predictor,
opacity cap, rasterizer, and photometric loss are held fixed, it asks whether the
same predicted surface samples are covered better when their covariance is
tangent-aligned and locally reorientable. The result is not a claim that the
reconstruction pipeline or metric depth improves; it is a covariance-parameterization
gain measured through rendering quality, re-rendering stability, and the
diagnostics below.

\subsection{Robustness to high-resolution re-rendering}
\label{sec:exp-robust}

Within the single-Gaussian-per-pixel group, PaveSplat has the best PSNR, SSIM,
and LPIPS at every scale, and this lead widens beyond $2\times$.
Table~\ref{tab:main} evaluates the same prediction under
high-resolution re-rendering: gaps that the rasterizer's one-pixel low-pass
filter~\cite{kerbl3dgs} covers at training resolution are exposed on larger grids,
following SurfSplat's
rendering-consistency argument~\cite{he2026surfsplat}. Against the same-recipe
analytic baseline,
PaveSplat degrades less on all three metrics: the $1\!\to\!4\times$
($1024^2$) PSNR drop is $\valCoreLargeHRRCdropPSNR$ vs.\
$\valBaseLargeHRRCdropPSNR$~dB, and at $8\times$ ($2048^2$) the gap is wider
still (PSNR \valCoreLargeHRRCeightPSNR{} vs.\ \valBaseLargeHRRCeightPSNR\,dB),
with smaller SSIM drop and LPIPS rise throughout; at $2\times$ it also leads on
all three metrics. The same ordering holds per scene: PaveSplat exceeds the
same-recipe baseline on
\psWinPSNRLargeFour/\psWinSSIMLargeFour/\psWinLPIPSLargeFour{} of scenes at
$4\times$ and \psWinPSNRLargeEight/\psWinSSIMLargeEight/\psWinLPIPSLargeEight{} at
$8\times$ (PSNR/SSIM/LPIPS; Holm-corrected within scale).

The protocol also sets the scope of the claim. HRRC reuses the same prediction
and changes only the rasterization grid, so it is a coverage stress test rather
than a replacement for native high-resolution capture. The per-resolution
PSNR/SSIM/LPIPS columns are correlated views of this stress test; we use them to
locate where support breaks, not as independent evidence streams.

Fig.~\ref{fig:qualitative} shows $8\times$ examples with the largest
$1\!\times\!\rightarrow\!8\times$ PSNR margin over the same-recipe baseline.
They illustrate, rather than quantify, the artifact summarized in
Table~\ref{tab:main}: native renders are visually comparable, but under
re-rendering the baselines lose coverage along thin structures such as door
frames, window panes, and fan blades.

\subsection{Zero-shot cross-dataset generalization}
\label{sec:exp-crossdata}

Table~\ref{tab:dl3dv} evaluates zero-shot transfer to DL3DV~\cite{ling2024dl3dv},
whose wide, real-world diversity stresses generalization beyond indoor
RealEstate10K. All models are trained on RE10K and evaluated under one protocol
at native and HRRC resolutions ($2\times$/$4\times$/$8\times$), with the
cost-volume baselines using their native scene normalization. PaveSplat leads
PSNR at every scale among single-Gaussian-per-pixel methods and ranks first or
second on SSIM/LPIPS; against the same-recipe baseline, the PSNR gap reaches
\valCoreDLfourPSNR{} vs.\ \valBaseDLfourPSNR\,dB at $4\times$ and
\valCoreDLeightPSNR{} vs.\ \valBaseDLeightPSNR\,dB at $8\times$. We read this as
near-domain transfer, not universal OOD robustness: the supplement shows transfer
to indoor ScanNet, while on far-domain outdoor ACID the same-recipe pair is
roughly tied at native resolution and PaveSplat falls behind under re-rendering.
This is consistent with the residual reorienting covariance where the
depth-derived frame is reliable, not replacing the underlying depth estimate. On
RE10K alone, we make no geometry-level claim; the evidence concerns rendering,
re-rendering, and transfer under the stated protocols.

\subsection{Mechanism analysis}
\label{sec:exp-mechanism}

\begin{figure}[t]
  \centering
  \includegraphics[width=0.95\linewidth]{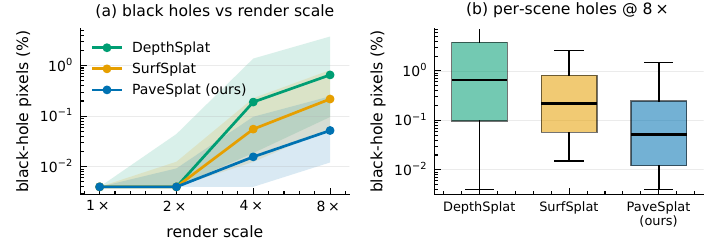}
  \caption{\textbf{PaveSplat leaves fewer coverage holes under
  high-resolution re-rendering} (RealEstate10K). The coverage-hole rate
  (black-hole fraction) $=$ share of pixels rendering near-black (max RGB channel
  ${<}0.02$) where the GT has content (max channel ${>}0.02$). \textbf{(a)}~vs.\ re-render scale
  (per-scene median line, IQR band, log-$y$): holes open under upsampling, fewest
  for PaveSplat at every re-rendered scale. \textbf{(b)}~per-scene distribution at $8\times$
  (box $=$ median/IQR, whiskers to the 10th/90th percentile): PaveSplat $<$
  SurfSplat $<$ DepthSplat (per-scene Wilcoxon, both $p{<}10^{-10}$).}
  \label{fig:coverage}
\end{figure}

\afterpage{\begin{table}[H]
  \centering
  \caption{\textbf{Mechanism checks} (means over the last 10K of the
  150K-step run), in three blocks: in-plane \emph{shape} (extent),
  \emph{orientation} (rotation residual $\dq$), and implemented-frame validity.
  n/a\,$=$\,undefined for the analytic baseline. These check the mechanism and
  numerical validity, not geometric accuracy; the residual is isolated in
  Tab.~\ref{tab:ablation}.}
  \label{tab:diagnostics}
  \begin{tabular}{@{}lcc@{}}
    \toprule
    Diagnostic & SurfSplat & PaveSplat (ours) \\
    \midrule
    \multicolumn{3}{@{}l}{\emph{In-plane shape (extent)}}\\
    \quad anisotropy $s_{\min}/s_{\max}$ (lower\,$=$\,more elongated) & \anisoBaseLargeLate & \anisoCoreLargeLate \\
    \quad near-isotropic fraction ($s_{\min}/s_{\max}{>}0.9$)        & \isoFracBaseLargeLate & \isoFracCoreLargeLate \\
    \quad scale-mult.\ floor saturation (lower\,$=$\,fewer shrunk)   & \scalesMulMinFracBaseLargeLate & \scalesMulMinFracCoreLargeLate \\
    \addlinespace
    \multicolumn{3}{@{}l}{\emph{Orientation (rotation residual $\dq$)}}\\
    \quad residual angle $\theta_\Delta$ mean (rad) & n/a & \dqAngleLargeLate~($\dqAngleLargeLateDeg^\circ$) \\
    \quad residual angle $\theta_\Delta$ p99 (rad)  & n/a & \dqPninetynineLargeLate~($\dqPninetynineLargeLateDeg^\circ$) \\
    \addlinespace
    \multicolumn{3}{@{}l}{\emph{Implemented-frame validity}}\\
    \quad orthonormality err.\ $\lVert R^{\!\top}\!R{-}I\rVert$ & $\frameOrthErrBaseLargeLate$ & $\frameOrthErrLargeLate$ \\
    \quad determinant (min) & \frameDetMinBaseLargeLate & \frameDetMinLargeLate \\
    \bottomrule
  \end{tabular}
\end{table}
}

Tab.~\ref{tab:diagnostics} reports \emph{mechanism checks}: logging-only
quantities that test whether each covariance aspect behaves as designed and
whether the implemented frame is numerically valid, not metrics of geometric
accuracy. The two covariance-correction aspects act on different parts of the
covariance, and we report them separately. The in-plane \emph{shape}
aspect keeps surfels from degenerating. Under the tangent-aligned frame the
surfels stay more elongated than under the normal-only analytic frame
($s_{\min}/s_{\max}$ \anisoCoreLargeLate{} vs.\ \anisoBaseLargeLate, with the
near-isotropic fraction \isoFracCoreLargeLate{} vs.\ \isoFracBaseLargeLate). With
the same bounded scale multipliers as the baseline, they also stay larger: the
analytic baseline pins \scalesMulMinFracBaseLargeLate{} of its in-plane
multipliers to the floor against \scalesMulMinFracCoreLargeLate{} for PaveSplat,
so its radii shrink toward the smallest scale while ours keep more in-plane
extent. One reading is that the correctable frame lets the network keep these
larger, elongated surfels rather than shrinking and rounding them.

The \emph{orientation} aspect then aims these surfels at the surface. The
rotation residual, summarized by its angle
$\theta_\Delta=2\arccos|\langle\dq\rangle_w|$, is used rather than collapsing to
identity (Tab.~\ref{tab:diagnostics}), with a mean angle of \dqAngleLargeLate\,rad
($\approx$\dqAngleLargeLateDeg$^\circ$) and a p99 of \dqPninetynineLargeLate\,rad
($\approx$\dqPninetynineLargeLateDeg$^\circ$). Spatially, the residual is
structured rather than diffuse, concentrating at depth discontinuities and
grazing surfaces, and the composed frame stays a numerically exact rotation
(Tab.~\ref{tab:diagnostics}).

These differences in surfel shape and extent show up directly as coverage. On
the black background an uncovered pixel renders black, so we measure the
coverage-hole rate (the black-hole fraction) --- the share of pixels that render
black where the ground truth has content (Fig.~\ref{fig:coverage}); read from the
final rendered image, it compares fairly across rasterizers. Holes are negligible
at native resolution and open under re-rendering, where the covariance regime
dominates and PaveSplat leaves the fewest: the gap widens with scale and holds per
scene, not only in the mean (per-scene Wilcoxon vs.\ the analytic baseline at
$8\times$, $p{<}10^{-10}$). These holes are a localized coverage failure ---
missing surface support along thin structures, not diffuse blur --- so leaving
fewer of them supports the reading that the high-resolution advantage comes from
maintaining surface support (Tab.~\ref{tab:main}).

We are careful about attribution: the in-plane shape and extent are set by the
scale multipliers and the tangent frame, not the residual, so the mechanism checks
show all three aspects move in the expected direction, not that the residual
alone drives the gain. The ablation in Tab.~\ref{tab:ablation} adds the frame and
then the residual along the construction path. Geometry-level claims additionally
require ground-truth depth: on zero-shot ScanNet, where ground-truth depth enables
rendered-normal evaluation, the same correction modestly improves normal alignment
--- evidence that the residual reorients surfels toward the surface rather than
re-estimating its depth (full protocol and numbers in the supplement).

\FloatBarrier
\subsection{Ablation study}
\label{sec:exp-ablation}

\begin{table}[t]
  \centering
  \caption{\textbf{PaveSplat design-path ablation on RealEstate10K} (identical
  recipe per row). Starting from the analytic baseline, we add the tangent frame
  and residual; columns report anisotropy, $8\times$ black-hole fraction, and
  $1\times$/$8\times$ PSNR.}
  \label{tab:ablation}
  \setlength{\tabcolsep}{6pt}
  \resizebox{\textwidth}{!}{%
  \begin{tabular}{@{}lcccccc@{}}
    \toprule
    Configuration & frame & $\dq$ & $s_{\min}/s_{\max}\!\downarrow$ & black\%$_{8\times}\!\downarrow$ & PSNR$_{1\times}\!\uparrow$ & PSNR$_{8\times}\!\uparrow$ \\
    \midrule
    SurfSplat$^{\ddagger}$                & normal  & --     & \anisoBaseLargeLate & \blackBaseLargeEight & \valBaseLargePSNR & \valBaseLargeHRRCeightPSNR \\
    \quad $+$ tangent frame              & tangent & --     & \anisoDoneLargeLate & \blackDoneLargeEight & \valDoneLargePSNR & \valDoneLargeHRRCeightPSNR \\
    \quad\quad $+$ residual $\dq$ (Ours) & tangent & \cmark & \anisoCoreLargeLate & \blackCoreLargeEight & \valCoreLargePSNR & \valCoreLargeHRRCeightPSNR \\
    \bottomrule
  \end{tabular}%
  }
\end{table}

Tab.~\ref{tab:ablation} ablates the construction of PaveSplat's covariance along
its design path, starting from the same-recipe SurfSplat baseline. Aligning the
in-plane frame with the measured surface tangents lowers the anisotropy ratio
$s_{\min}/s_{\max}$ from \anisoBaseLargeLate{} to \anisoDoneLargeLate{}, elongating
the surfels along the surface; this alone, however, does not resolve the
high-resolution coverage failure: the tangent frame supplies anisotropic support but
can still aim a surfel off the surface, and the mean black-hole fraction at $8\times$
rises from \blackBaseLargeEight\% to \blackDoneLargeEight\%. This rise is a thin
tail: 64 of the \nScenes{} test scenes carry the entire increase, while 59.2\% of
scenes have fewer holes than under the baseline (per-scene Wilcoxon,
$p{<}10^{-10}$). This pattern is consistent with an elongated axis aimed off the
surface at the depth discontinuities and grazing surfaces where the residual
concentrates (Sec.~\ref{sec:exp-mechanism}). Adding the identity-initialized residual
reorientation to this prior is what delivers the high-resolution
robustness: it lowers the black-hole fraction to \blackCoreLargeEight\%, below the
baseline, and raises $8\times$ PSNR from the tangent-only \valDoneLargeHRRCeightPSNR{}
to \valCoreLargeHRRCeightPSNR{}, while keeping the frame's anisotropy. This step
isolates the residual: it is the only change between the two tangent-frame rows,
whereas the first step also switches from the released construction to our
camera-space formulation (Sec.~\ref{sec:method-frame}), so the table does not
establish that the frame and the residual are independently beneficial. The
residual is our only added parameter, and the full model
stays the strongest single-Gaussian method at every scale (Tab.~\ref{tab:main}).
\FloatBarrier

\section{Conclusion and Limitations}
\label{sec:conclusion}
\label{sec:limitations}

We present PaveSplat, a residually corrected surface-aligned covariance for
feed-forward Gaussian splatting. It keeps SurfSplat's analytic covariance as the
operating point, aligns the in-plane frame with measured tangents, maps the
covariance once to world space, and composes an identity-initialized quaternion
residual that reorients each surfel locally. Under a controlled RealEstate10K
recipe, PaveSplat improves over the same-recipe baseline, leads single-Gaussian-per-pixel methods
at every scale, degrades less under re-rendering, and leaves fewer coverage
holes. Diagnostics and
ablation show the residual moves away from identity and contributes after the
tangent-aligned frame.

The scope is limited. HRRC is one coverage stress test on re-rendered
predictions, not a replacement for native high-resolution capture or independent
signals; the ablation isolates the residual on one backbone and dataset, not
every decomposition. The residual adapts orientation only: zero-shot ScanNet
shows a modest rendered-normal gain, not a metric-depth claim. Among the
reported zero-shot settings the benefit holds on near-domain DL3DV and ScanNet
but not far-domain ACID (tied at native resolution, worse re-rendered); the
residual costs \runtimeOverheadPct{} encoder time.

\subsubsection*{\ackname}
This work was supported by Electronics and Telecommunications Research Institute
(ETRI) grant funded by the Korean government [26ZC1100, Development of
Spatial Media Technology and Interaction Technology for Convergence of the Real
and Virtual World].

\bibliographystyle{splncs04}
\bibliography{bib/refs}

\end{document}